\documentclass{article}

\usepackage[a4paper, top=2cm, bottom=2cm, left=3cm, right=3cm]{geometry}

\usepackage{amsmath}
\usepackage{graphicx}
\usepackage[colorlinks=true, allcolors=blue]{hyperref}
\usepackage{subcaption} % 子图
\usepackage{enumitem} % 列表控制
\usepackage{multirow} % 表格多行
\usepackage{booktabs} % 优化表格排版，创建三线表
\usepackage{xcolor} % 颜色支持
\usepackage{abstract} % 摘要格式控制
\usepackage{url} % 处理URL
\usepackage{caption} % 允许 captionof
\usepackage{float} % 可以使用H，但建议让latex自动处理
\usepackage{tabularx} % 用于创建自动调整列宽的表格
\usepackage{geometry}
\usepackage{array}
\usepackage{ragged2e}

\newcolumntype{Y}[1]{>{\RaggedRight\hsize=#1\hsize}X}

\title{\textbf{MM-Food-100K: A 100,000-Sample Multimodal Food Intelligence Dataset with Verifiable Provenance}}
\author{
    Yi Dong\thanks{sky@inductive.network, Codatta} \and
    Yusuke Muraoka\thanks{yusuke.muraoka@gokite.ai, Kite AI} \and
    Scott Shi\thanks{Scott.shi@gokite.ai, Kite AI} \and
    Yi Zhang\thanks{PhD, yi@inductive.network, Codatta}
}
\date{
    Binance Wallet Community\\
    Codatta Community
}

\begin{document}

\maketitle

\begin{abstract}
We present \textbf{MM-Food-100K}, a high-quality, 100,000-sample multimodal food intelligence dataset designed for fine-tuning AI models. Our proposed data sourcing workflow combines community contributions with automated quality review by Large Vision-Language Models, resulting in real-world photos with rich, multi-level annotations. We demonstrate the dataset's utility by showing that models fine-tuned on MM-Food-100K consistently outperform original Large Vision-Language Models on food classification and regression tasks. We also introduce Codatta Protocol, a novel data contribution framework that uses a privacy-preserving blockchain protocol to track data provenance and enable royalty-based rewards for contributors. MM-Food-100K is released for public use, and the Codatta Protocol provides a new paradigm for building and sustaining high-quality, community-sourced datasets.
\end{abstract}

\section{Introduction}
Human-generated data remains indispensable for training reliable multimodal AI systems. While synthetic data can amplify scale, it tends to recycle model priors, misses edge-case realism, and struggles with grounded facts. As a result, real-world sourced data and advanced, knowledge-rich labeling continues to be in high demand. Today’s human-intelligence pipelines exhibit a \textbf{dual-market failure}. On the demand side, businesses face high costs to obtain domain-rich, verifiable data. On the supply side, knowledge providers are frequently underpaid or misaligned with downstream value, while attribution is rarely preserved, depressing participation and diversity of expertise.

\begin{figure}[htbp]
    \centering
    \includegraphics[width=1\textwidth]{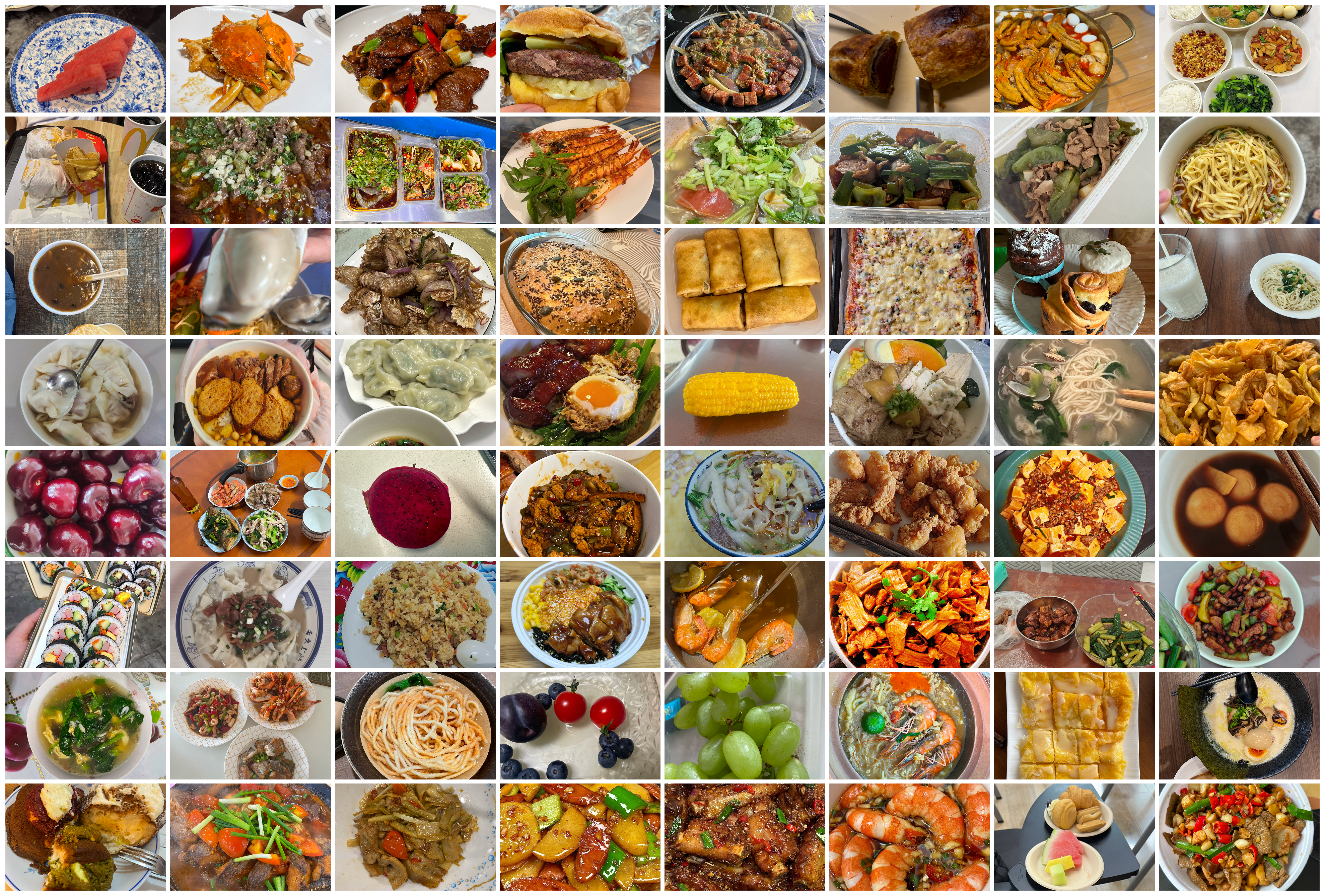}
    \caption{Overview of Food Intelligence Dataset Samples: A collage of diverse dishes and ingredients from multiple cuisines, showcasing the variety in the Food Intelligence dataset-including noodles, dumplings, soups, stir-fries, seafood, meats, breads, fruits, and snacks-captured in varied lighting, perspectives, and serving styles.}
    \label{fig:example}
\end{figure}

In this paper we introduce \textbf{MM-Food-100K}, a 100,000-sample, multimodal (image + textual annotation) dataset released for public research. The dataset samples are diverse, showcasing a variety of dishes and ingredients, as shown in Figure \ref{fig:example}. MM-Food-100K is a curated $\sim$10\% open subset of an original \textbf{1.2-million}, quality-accepted corpus collected in six weeks from 87,000+ contributors. Data is sourced via a partnership campaign between \textbf{Binance Wallet} and the \textbf{Codatta protocol}, which combines community sourcing with configurable AI-assisted quality checks. Each submission is linked to an on-chain \textbf{wallet address} to provide verifiable provenance and attribution; a fully on-chain protocol is on the roadmap. Importantly, contributors were invited with zero upfront payment; instead, the model emphasizes attribution and prospective revenue sharing for the retained commercial portion of the corpus, aligning incentives without front-loading cost or risk.

\textbf{Contributions:}
\begin{itemize}
    \item \textbf{Dataset (primary)}. We release \textbf{MM-Food-100K}, a public multimodal food-intelligence dataset annotated for dish/cuisine, visible ingredients, portion and scale cues and etc., with evidence and provenance for auditability. We demonstrate its utility by fine-tuning large vision–language models for image-based food-related prediction (such as dish recognition), observing consistent improvements over out-of-box baselines.
    \item \textbf{Framework (secondary)}. We describe briefly the \textbf{Codatta} blockchain-based data contribution model—data infrastructure for privacy-preserving transparency, structured task/schema design, and configurable human/AI quality assurance, contributor profile and reputation system, and wallet-linked provenance —that enabled verifiable data acquisition at scale and royalty-based payment business model via data classification.
\end{itemize}

\section{The MM-Food-100K Dataset}
\subsection{Overview and applications}
\textbf{MM-Food-100K} is a public, 100,000-sample subset of a larger $\sim$1.2M food corpus curated for \textbf{fine-grained food intelligence}. Each record pairs an image with structured annotations that enable recognition (dish/cuisine), ingredient extraction, portion and calorie estimation, cooking-method reasoning, and authenticity checks. The dataset targets practical applications such as image-based nutrition estimation, recipe recommendation, diet logging/health monitoring, and fine-grained vision research.

\textbf{Why another dataset?} A scan of publicly available food datasets highlights three recurring gaps: (i) insufficient diversity and scale, (ii) limited annotation, “monolithic” labels with little schema depth (e.g., no portion/nutrition), and (iii) curated imagery that diverges from real-world photos. MM-Food-100K addresses these limitations by combining breadth with \textbf{multi-level, structured annotations} tailored to downstream AI tasks.

\subsection{Schema (what a record contains)}
\begin{figure}[htbp]
    \centering
    \includegraphics[width=1\linewidth]{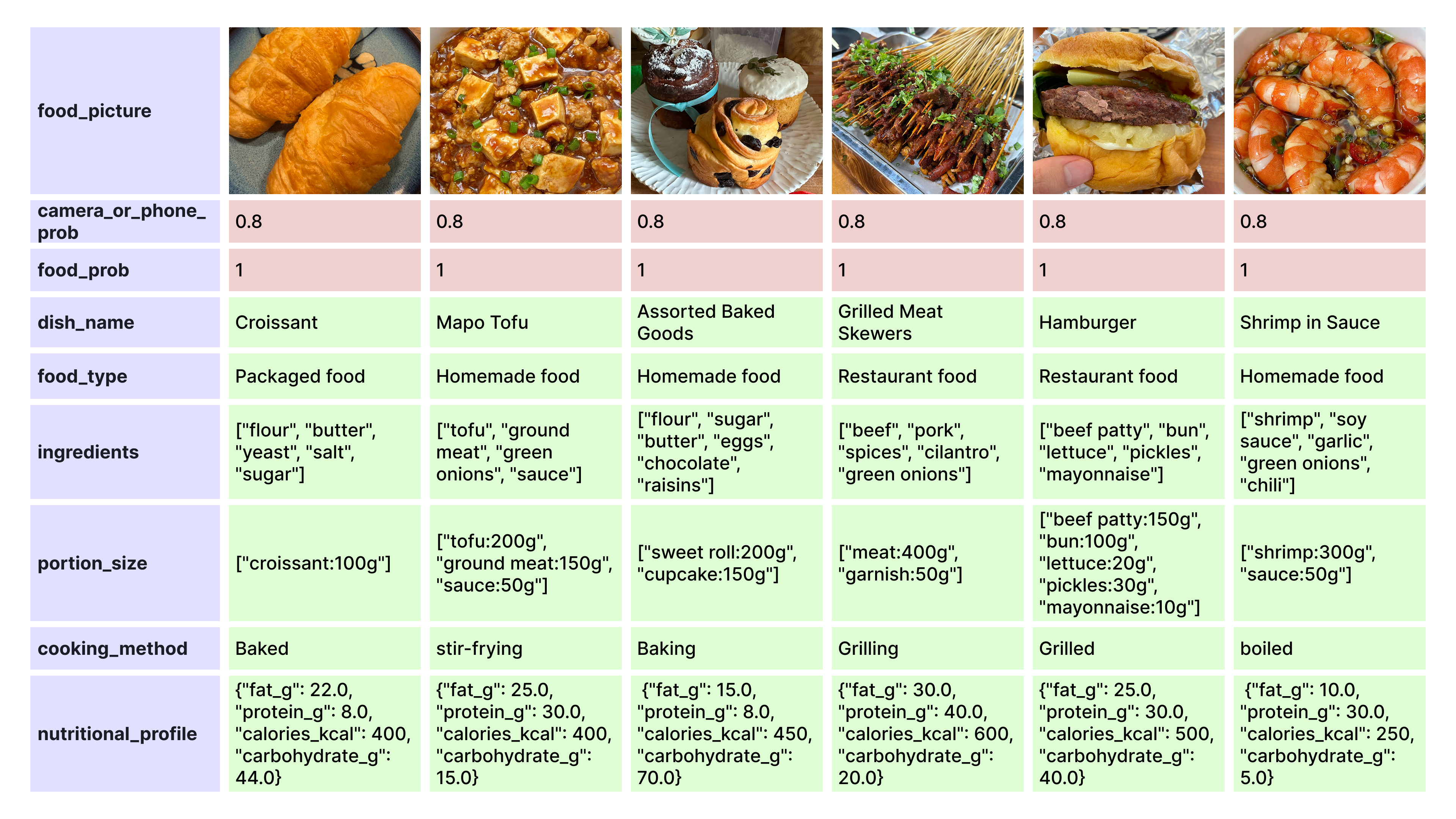}
    \caption{Selected Samples from the Food Intelligence Dataset: Images with Mixed Human and AI Annotations. Examples of food images with mixed annotations—green cells are human-labeled, red cells are AI-predicted—showing dish details, ingredients, cooking methods, and nutrition.}
    \label{fig:fig2}
\end{figure}

Each item includes an image link and a JSON metadata block with the following key fields (non-exhaustive): \texttt{dish\_name}, \texttt{food\_type} (homemade/restaurant/packaged/raw), \texttt{ingredients}, \texttt{portion\_size} (ingredient-level estimates), \texttt{nutritional\_profile} (kcal/protein/fat/carbs when available), \texttt{cooking\_method}, and authenticity indicators (\texttt{camera\_or\_phone\_prob}, \texttt{online\_download\_prob}, \texttt{food\_prob}). A visual representation of these annotations, which can be a mix of human and AI labels, is provided in Figure \ref{fig:fig2}, and the core schema fields are detailed in Table \ref{tab:schema}. This design supports multi-task learning (classification, extraction, and regression) without changing file formats.

\begin{table}[htbp]
    \centering
    \caption{Core schema fields}
    \label{tab:schema}
    \begin{tabular}{llp{5cm}}
        \toprule
        \textbf{Group} & \textbf{Example fields} & \textbf{Purpose} \\
        \midrule
        Identity & dish\_name, food\_type, cuisine tag & recognition / taxonomy \\
        Ingredients & ingredients[] & extraction / reasoning \\
        Portion & portion\_size (g/ml per item) & portion / calorie models \\
        Nutrition & nutritional\_profile (kcal, P/F/C) & nutrition prediction \\
        Preparation & cooking\_method & method reasoning \\
        Authenticity & camera\_or\_phone\_prob, online\_download\_prob, food\_prob & data realism filters \\
        \bottomrule
    \end{tabular}
\end{table}

\subsection{Codatta's Two-Stage, AI-Augmented Data Sourcing Workflow}
\begin{figure}[htbp]
    \centering
    \includegraphics[width=0.8\linewidth]{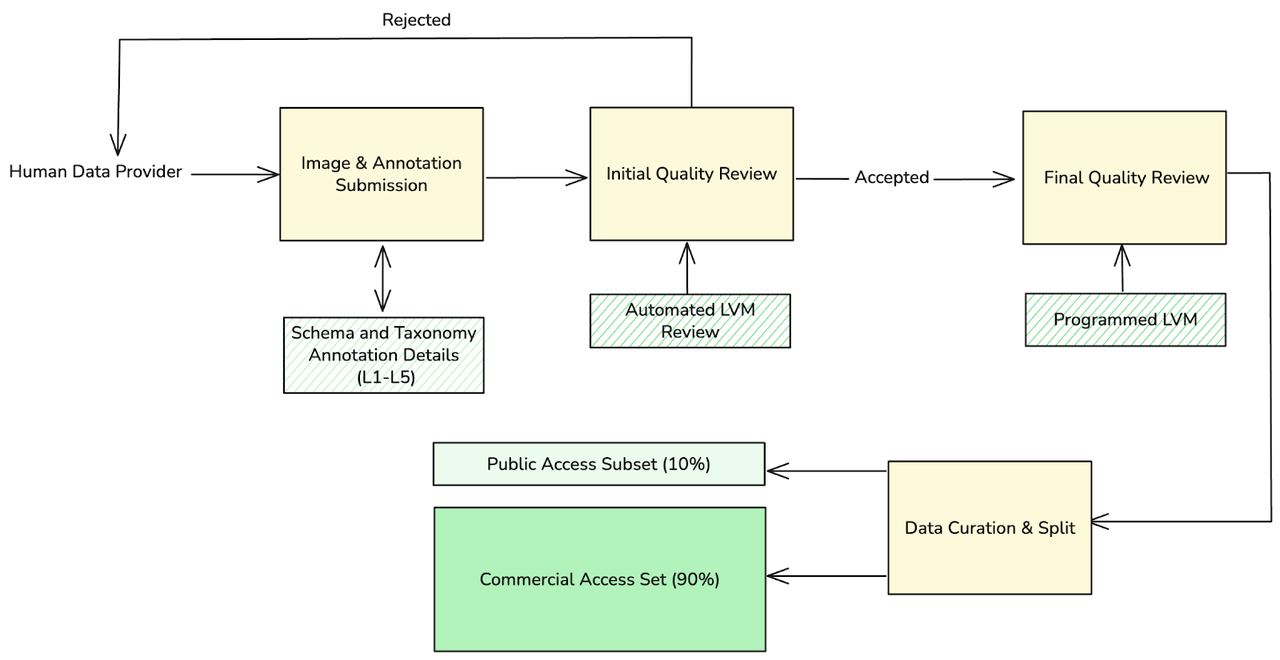}
    \caption{Data Pipeline: Two-Stage Quality Review and Dual-Access Distribution. A workflow for reviewing food image–annotation pairs with quick initial checks, detailed final reviews, and a split into public and commercial sets to balance openness and provider rewards.}
    \label{fig:pipeline}
\end{figure}

Creating high-quality, large-scale multimodal datasets from diverse community sources is challenging due to the trade-off between quality assurance and operational costs. To address this, we designed and implemented a novel two-stage, AI-augmented data sourcing workflow. As illustrated in Figure \ref{fig:pipeline}, this pipeline is a core innovation of the Codatta protocol, turning heterogeneous user-generated content into verifiable, high-fidelity records at scale. Our method consists of the following key steps:
\begin{enumerate}[label=(\arabic*)]
    \item \textbf{Submission and Progressive Enrichment}: Initially, a \textbf{Human Data Provider} submits an image and corresponding annotations. These annotations follow a structured schema and taxonomy (L1-L5), allowing for progressive enrichment from basic dish names to complex, evidence-backed nutritional and portion cues.
    \item \textbf{Initial Quality Review}: Each submission proceeds to an \textbf{Initial Quality Review} stage, where an \textbf{Automated LVM Review} conducts a rapid vetting process. This first-pass check filters out low-quality or irrelevant submissions and provides near-real-time feedback. If a submission is rejected, the data provider can revise and resubmit.
    \item \textbf{Final Quality Review}: Submissions that pass the initial review are forwarded to a more thorough \textbf{Final Quality Review}. This stage leverages a more sophisticated, \textbf{Programmed LVM} to perform detailed checks, ensuring the accuracy and integrity of the data.
    \item \textbf{Data Curation and Distribution}: Accepted and validated data is then processed for \textbf{Data Curation} and deduplication. It is subsequently split into two distinct sets: a \textbf{Public Access Subset (10\%)} for the research community and a \textbf{Commercial Access Subset (90\%)} for commercial use. This dual-access distribution model retains wallet-linked provenance, laying the foundation for future royalty-based rewards.
\end{enumerate}
This two-stage, AI-augmented workflow achieves high-quality data acquisition at scale and a sustainable balance between cost and data quality. It establishes a transparent process and a reward mechanism that benefits both community contributors and downstream applications.

\subsection{Descriptive statistics (what the open subset looks like)}
The dataset card reports \textbf{100,000} images with a broad mix of source contexts (`food\_type'), illustrating real-world capture conditions and variety. The distribution of `food\_type' is as follows: Homemade (46,555), Restaurant (35,461), Raw vegetables \& fruits (9,357), Packaged (8,354), and Others (273). The card also includes authenticity distributions (e.g., \texttt{camera\_or\_phone\_prob} bins at 0.8 and 0.7 totaling $\sim$99k images), underscoring the emphasis on user-captured photos. A snapshot of these distributions is presented in Table \ref{fig:table-snapshot}. For a complete overview, see the full dataset card on \href{https://huggingface.co/datasets/Codatta/MM-Food-100K/blob/main/README.md}{Hugging Face}.

\begin{table}[htbp]
    \centering
    \caption{Snapshot of reported distribution.}
    \label{fig:table-snapshot}
    \begin{tabular}{lll}
        \toprule
        \textbf{Dimension} & \textbf{Values (examples)} & \textbf{Counts} \\
        \midrule
        Food type & Homemade / Restaurant / Raw / & 46,555 / 35,461 / 9,357 / \\
        & Packaged / Other & 8,354 / 273 \\
        \addlinespace
        Authenticity & 0.9 / 0.85 / 0.8 / 0.7 / 0.6 & 200 / 161 / 47,879 / \\
        (camera\_or\_phone\_prob) & & 51,629 / 131 \\
        \bottomrule
    \end{tabular}
\end{table}

\section{Codatta Protocol}
\subsection{Challenges of legacy data markets}
Knowledge-rich labeling and real-world data sourcing are expensive to acquire and in high-demand for improving foundational language models. Buyers face high and uncertain upfront costs, opaque provenance, uneven quality assurance and limited auditability or compliance guarantees. Contributors receive relatively low flat fees, forfeit attribution, and capture no upside from downstream reuse, which depresses long-term participation, skill development and discouragement for skillful experts to take part-time contribution tasks. Operationally, static rule checks police formatting rather than truthfulness, evidence is rarely captured in a reusable form, and usage cannot be reliably traced once data leave the platform.

\subsection{Solution: Royalty-based Payment for Data Access.}
\begin{figure}[htbp]
    \centering
    \includegraphics[width=0.8\linewidth]{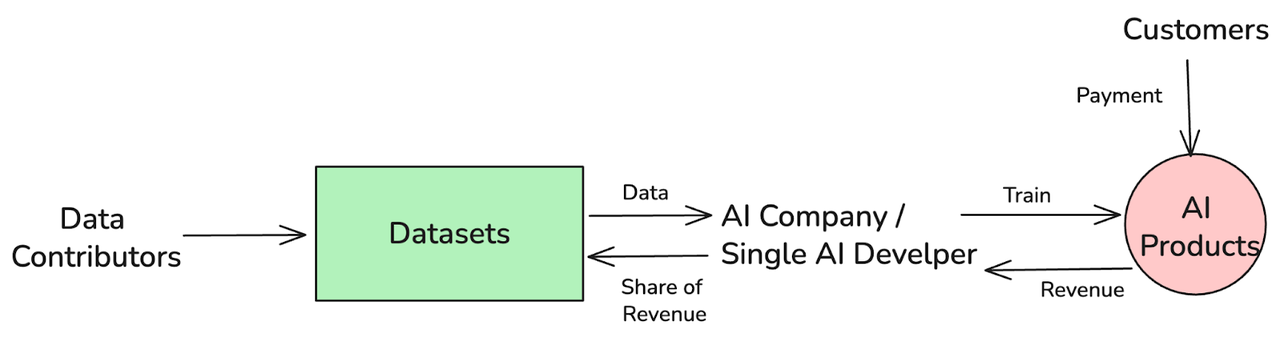}
    \caption{Royalty-Based Payment Flow in the Codatta Protocol. Usage-linked revenue sharing from AI products back to data contributors (wallet-linked provenance; quality/usage-weighted payouts).}
    \label{fig:royalty-flow}
\end{figure}

In Codatta, royalties link contributor rewards to realized product value. Data contributors supply items that enter a curated dataset; AI developers or companies license and train on this dataset to build models and applications. When customers pay for products derived from the dataset—directly through licenses or indirectly through model use—a pre-agreed share of that revenue flows back to the dataset and is distributed to contributors or data owners according to provenance records and quality-/usage-weighted splits. This arrangement reverses the one-off, up-front payment paradigm: buyers assume less risk because costs scale with actual use, while contributors retain ongoing upside as their data continue to power deployed AI products. Figure \ref{fig:royalty-flow} illustrates this royalty-based payment flow.

Economically, the royalty model converts data acquisition from a large, upfront capital expense into a usage-based operating expense: buyers pay only as models trained on the dataset generate value. This lowers prior risk, allows small pilots before scale, and ties spend to actual adoption and performance rather than speculative volume. For contributors, recurring royalties reward durable quality and sustained participation, creating a feedback loop in which better data lead to better products and, in turn, higher payouts distributed via wallet-linked provenance.

\subsection{Privacy-preserving transparency for data assetification}
\begin{figure}[htbp]
    \centering
    \includegraphics[width=0.7\linewidth]{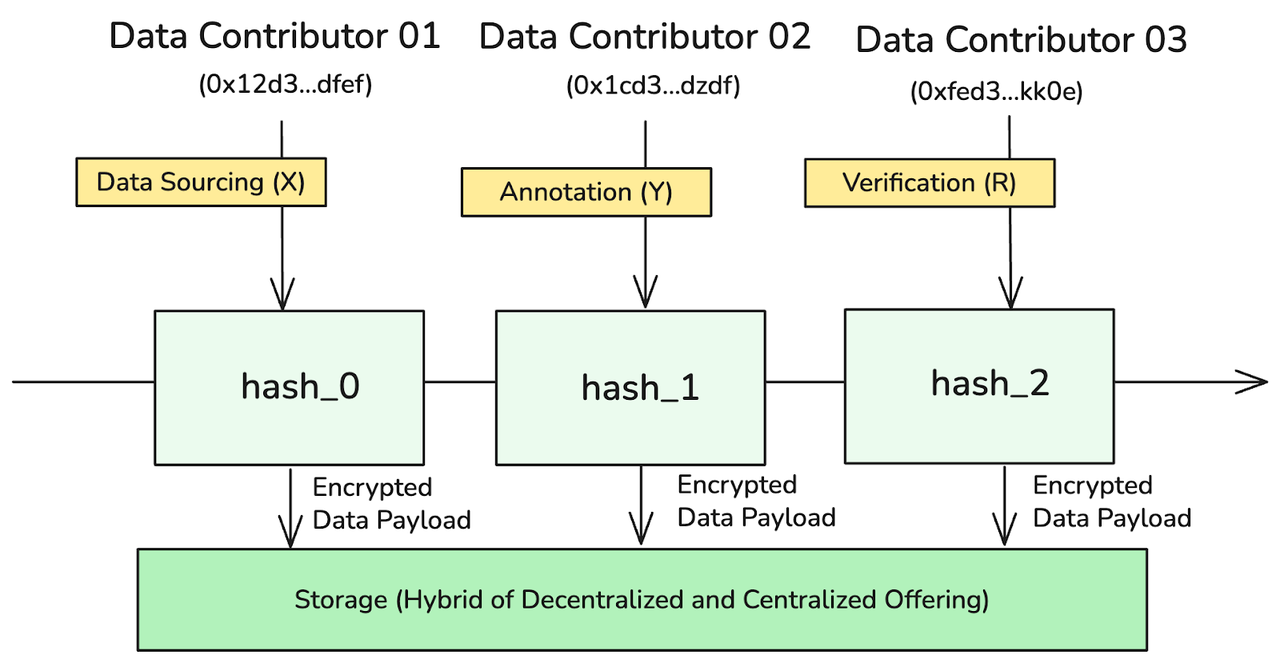}
    \caption{Blockchain-Based Data Contribution Ledger with Privacy-Preserving Transparency. A decentralized attestation layer anchors sourcing, annotation, verification, and adoption events as immutable hashed records, while encrypted payloads are managed in a hybrid decentralized–centralized storage architecture. This preserves verifiable provenance for attribution and royalties, without exposing sensitive data.}
    \label{fig:ledger}
\end{figure}

A royalty-based market only works if provenance and usage can be trusted without exposing the underlying data. To meet this requirement, we treat each contributed record as a traceable digital asset with two complementary properties: public \textbf{attestability} of its existence, authorship, and evolution; and \textbf{confidentiality} of its payload for research or commercial use. Attestability enables fair attribution and accurate royalty allocation; confidentiality preserves contributor privacy and protects the economic value of the dataset.

Our design employs a blockchain-anchored contribution ledger, as depicted in Figure \ref{fig:ledger}. Each sourcing, annotation, verification, and adoption event is recorded in a secure ledger tied to contributor wallets and time-stamped. Periodically, compact cryptographic commitments to batches of these events are anchored to a public chain, making the history tamper-evident and independently checkable without placing raw data on-chain. This provides non-repudiation—contributors can prove inclusion and credit—while keeping operational cost and sensitive content off the ledger.

Payloads (images, evidence screenshots, review notes) are stored in a hybrid architecture. Publicly verifiable metadata—content fingerprints, dataset/version identifiers, role credits—remain visible; the corresponding files are encrypted and gated behind controlled access for research or licensed use. Selective disclosure allows auditors or partners to verify specific claims (for example, that a nutrition label supports a calorie value) without broad data release. The result is \textbf{privacy-preserving transparency}: anyone can verify who contributed what and when, while only authorized parties can view the data itself.

This infrastructure enables \textbf{assetification} of the corpus. Because each record has a stable identity, provenance, and verifiable usage trail, royalties can be computed from observed adoption (in releases, training runs, or licenses) and distributed to credited wallets. At the same time, contributors retain public attribution for their work, and buyers gain an auditable chain of custody for compliance. Practically, we are rolling this out iteratively: today’s system uses verifiable off-chain records with periodic on-chain attestations; our ongoing “Booster” campaign extends this by publishing an immutable, publicly checkable list of contributions to honor ownership at scale. As the attestation process matures, more of the accounting can be automated, but the core guarantees—traceability, integrity, and controlled access—remain the backbone of the business model.

\subsection{Implementation Status and Commercialization Plan}
The decentralization of the data infrastructure is in active development rather than fully realized. Our present system anchors provenance and usage with verifiable off-chain records and periodic public attestations; the end-state—contract-mediated revenue pools, on-chain distribution, and community governance—remains on the roadmap. The recent Binance Wallet campaign should be read as a short-term accelerator to stress-test the royalty concept at scale: it provided wallet-linked onboarding, fast growth in contributions, and an initial public basis for attributing ownership without exposing underlying data.

As the dataset transitions from research release to commercial use, royalty sharing will be executed against the contributor registry established during the campaign. In practice, commercialization events will fund a payout pool, and distributions will reference the recorded contributor addresses and role credits. Public attestations will document which contributions were included, while detailed payloads remain access-controlled to preserve privacy and commercial value. This approach preserves the core guarantees—traceable attribution, auditable usage, and fair reward—while allowing the decentralization stack to mature incrementally.

\section{Experiments}
\subsection{Overview and setup (quick study)}
We run a minimal, decision-oriented experiment to obtain an early signal of data value. The primary task is \textbf{image-based nutrition prediction} (kilocalories). We compare two foundation models—\textbf{ChatGPT-4o} and \textbf{Qwen-Max}—in their \textbf{out-of-box} form and after supervised fine-tuning (\textbf{SFT}) on \textbf{MM-Food-100K}. Data are split \textbf{80/10/10} (train/validation/test), stratified by cuisine and source type (packaged, chain restaurant, homemade/street); near-duplicates are removed at ingest via perceptual hashing. Schema and check-logic versions associated with each item are fixed to a frozen release index for reproducibility.

The fine-tuning process is intentionally lightweight. For SFT, we use a simple regression head on the shared vision-language backbone. We optimize a Huber/MSE objective with mixed precision, fixed augmentations, and early stopping based on validation MAE. To attribute all performance differences to the data itself, not optimization variables, we keep seeds, batch sizes, and learning rates constant across all runs. The test split is identical for all models.

We design our evaluation to be task-specific:
\begin{itemize}
    \item \textbf{Dish name, ingredients, and cooking method}: We use a text-similarity-based win rate. In each comparison between two models (e.g., base vs. fine-tuned), we report the percentage of test cases where one model's output is judged to be more accurate or comprehensive.
    \item \textbf{Kilocalories}: We use standard regression metrics: \textbf{MAE (Mean Absolute Error)}, \textbf{RMSE (Root Mean Square Error)}, and \textbf{R$^2$}.
\end{itemize}

\subsection{Results}
The following tables present our findings, illustrating a clear pattern: SFT-ChatGPT-4o consistently outperforms all others, followed by the statistically similar ChatGPT-4o base and SFT-Qwen-Max, with Qwen-Max base trailing behind. The win rate comparison for categorical predictions is shown in Table \ref{tab:win-rate}, and the regression metrics for kilocalorie prediction are detailed in Table \ref{tab:regression-metrics}.

\begin{table}[htbp]
    \centering
    \caption{Win Rate Comparison for Categorical Predictions}
    \label{tab:win-rate}
    \begin{tabular}{l|ccc}
        \toprule
        \textbf{Model} & \textbf{Dish Name} & \textbf{Ingredients} & \textbf{Cooking Method} \\
        & \textbf{(Win Rate)} & \textbf{(Win Rate)} & \textbf{(Win Rate)} \\
        \midrule
        Qwen-Max (Base) & 42.1\% & 39.8\% & 41.3\% \\
        Qwen-Max (SFT) & 57.9\% & 60.2\% & 58.7\% \\
        \midrule
        GPT-4o (Base) & 48.9\% & 48.6\% & 48.8\% \\
        GPT-4o (SFT) & 51.1\% & 51.4\% & 51.2\% \\
        \bottomrule
    \end{tabular}
\end{table}

\begin{table}[htbp]
    \centering
    \caption{Regression Metrics for Kilocalorie Prediction}
    \label{tab:regression-metrics}
    \begin{tabular}{l|ccc}
        \toprule
        \textbf{Model} & \textbf{MAE (kcal) $\downarrow$} & \textbf{RMSE (kcal) $\downarrow$} & \textbf{R$^2$ $\uparrow$} \\
        \midrule
        Qwen-Max (Base) & 126.5 & 185.3 & 0.521 \\
        Qwen-Max (SFT) & 104.2 & 154.5 & 0.638 \\
        \midrule
        GPT-4o (Base) & 98.7 & 148.1 & 0.685 \\
        GPT-4o (SFT) & 95.8 & 144.3 & 0.706 \\
        \bottomrule
    \end{tabular}
\end{table}

\textbf{Observations.} Fine-tuning on MM-Food-100K significantly improves performance. For Qwen-Max, fine-tuning reduces calorie \textbf{MAE by approximately 17.6\%} (126.5 $\to$ 104.2 kcal). The \textbf{RMSE} sees a larger reduction of \textbf{16.6\%} (185.3 $\to$ 154.5 kcal), and the \textbf{R$^2$} value \textbf{improves by 22.4\%} (0.521 $\to$ 0.638). This pattern of improvement across the metrics demonstrates the dataset's value in enhancing predictive accuracy and reducing large errors. The win rates for dish name, ingredients, and cooking method also show substantial gains of 15.8, 20.4, and 17.4 percentage points, respectively. For ChatGPT-4o, the gains are smaller but still notable: calorie MAE drops by $\sim$2.9\% (98.7 $\to$ 95.8 kcal), RMSE by $\sim$2.6\% (148.1 $\to$ 144.3 kcal), and R$^2$ improves by $\sim$3.1\% (0.685 $\to$ 0.706). We observe the largest relative gains on homemade/street items, where the dataset's structured portion cues (Level 4) and evidence-backed nutrition (Level 2) provide the most valuable additional signal.

\subsection{Discussion}
These results support the central claim that \textbf{human, evidence-linked supervision} provides value beyond generic pretraining. The improvement pattern is consistent across two model families and persists when optimization is held constant, indicating that the gains derive from the data rather than from training tricks. The rough parity between ChatGPT-4o (base) and SFT-Qwen-Max underscores a practical point: access to a strong foundation model is helpful, but targeted supervision can lift a weaker base to comparable accuracy on domain tasks. Conversely, SFT-ChatGPT-4o achieves the best overall performance, suggesting that high-capacity models benefit most from schema-rich supervision.

This is an intentionally minimal study designed to deliver a fast, actionable signal. Next iterations will separate task-specific heads for dish/cuisine, ingredients, and portion; report per-domain strata (packaged, chain, homemade/street); and expand to additional model families and training sizes to characterize scaling. We will also link intake quality to outcomes by weighting training examples with the dataset’s QA scores. All future phases will reuse the same frozen indices and configuration templates to keep comparisons clean and auditable.

\section{Discussion and Future Work}

\textbf{Data value and limits.} MM-Food-100K delivers clear gains on image-based nutrition prediction and improves related signals (dish accuracy, portion error). The benefit comes from evidence-linked fields—nutrition proof and portion/scale cues—that generic pretraining lacks. Limits remain: homemade/street portions and calories carry uncertainty; some cuisines are under-represented; menus drift seasonally; filled-rates vary by field. Our release balances openness and incentives: 100K is public for research; the retained corpus is licensable with revenue-sharing to contributors, converting data spend from a large upfront cost to usage-based payment.

\textbf{Protocol and royalties.} The Codatta protocol targets the dual-market failure of legacy platforms by paying for \textbf{adoption}, not just task completion. Wallet-linked provenance and quality/usage-weighted splits connect data value to rewards. Implementation is iterative: accounting runs off-chain with public attestations today; the Binance Wallet campaign established a contributor registry that will be honored during commercialization; fuller decentralization (contract-based pools and automated disbursement) is on the roadmap.

\textbf{Next studies and the human-AI loop.} This quick study is a first signal. We will extend to dish/cuisine, portion, and ingredients with task-specific heads; add more model families and training sizes; and run ablations on schema depth and QA-weighted training, all on fixed splits for clean comparisons. Weekly instrumentation (A, AC, AD, AC/A, AD/AC, review time, cost/AD) guides updates to pre-accept gates and reviewer prompts: AI absorbs routine errors and raises first-pass acceptance; human audit focuses on edge cases. This interaction moves the cost–quality frontier outward and provides leading indicators for downstream model gains.

\section*{Appendix}

\subsection*{Schema and taxonomy-design rationale}
The schema and taxonomy are designed to turn heterogeneous food photos into \textbf{verifiable, model-ready supervision} while keeping collection costs predictable and quality auditable.

\subsection*{Design goals}
\begin{itemize}
    \item \textbf{Verifiability first:} every factual claim (e.g., calories, portion) links to \textbf{evidence} (label/menu/URL/receipt) and to \textbf{provenance} (wallet-linked contributor record).
    \item \textbf{Progressive enrichment:} annotations are organized into \textbf{five levels} so contributors (or AI agents) can start with low-friction L1 submissions and \textbf{upgrade} items to richer L2--L5 records.
    \item \textbf{Hybrid QA:} the same fields support \textbf{human}, \textbf{AI}, or \textbf{hybrid} quality checks (e.g., OCR validation, outlier flags, consensus sampling).
\end{itemize}

\subsection*{Why three domains}
We partition sources into \textbf{Packaged}, \textbf{Chain restaurant}, and \textbf{Homemade/Street} to align evidence requirements and QA cost with a \textbf{ground-truth gradient} and a \textbf{variance gradient}. Packaged items have printed nutrition (high verifiability, low visual variance); chain items have menu nutrition (medium verifiability); homemade/street items rely on estimates and portion cues (lower external ground truth, higher variance). This split lets us tune evidence rules and sampling rates \textbf{per domain}.

\subsection*{Why five levels}
Levels reflect \textbf{increasing effort and model utility}:
\begin{itemize}
    \item \textbf{L1}: image + name (seed the corpus).
    \item \textbf{L2}: nutrition with evidence (make records verifiable).
    \item \textbf{L3}: visible ingredients (enable extraction).
    \item \textbf{L4}: portion \& geometry (enable calorie/serving regression).
    \item \textbf{L5}: per-ingredient localization (enable detection/segmentation).
\end{itemize}

\begin{table}[htbp]
    \centering
    \small % 缩小字体
    \caption{Where the data comes from (source types)}
    \label{tab:data-sources}
    \begin{tabularx}{\textwidth}{l >{\RaggedRight}X >{\RaggedRight}X >{\RaggedRight}X >{\RaggedRight}X >{\RaggedRight}X >{\RaggedRight}X >{\RaggedRight}X}
        \toprule
        \textbf{Source type} & \textbf{Where it comes from} & \textbf{Photos to include} & \textbf{Nutrition source} & \textbf{Minimum detail level} & \textbf{Proof to attach} & \textbf{Default quality check} & \textbf{Common pitfalls} \\
        \midrule
        Packaged food & supermarket packs, barcode pages & front + back / label & per serving / per pack & 2 & label photo or clear OCR & auto checks + quick human review & duplicate SKUs; old labels \\
        \midrule
        Chain restaurant & brand website/menu, in-store boards & dish photo + menu screenshot & per serving / menu size & 2 & screenshot / URL with item marked & OCR match + human confirm & menu changes; seasonal items \\
        \midrule
        Homemade / street & personal photos, chef notes, receipts & dish photo & estimate / per 100g & 3 & short note or receipt (if any) & human-led review + model assist & portion bias; privacy; mixed plates \\
        \bottomrule
    \end{tabularx}
\end{table}

\begin{table}[htbp]
    \centering
    \small
    \caption{What each annotation level adds}
    \label{tab:data-sources-part2}
    \begin{tabularx}{\textwidth}{l >{\RaggedRight}X >{\RaggedRight}X >{\RaggedRight}X >{\RaggedRight}X >{\RaggedRight}X}
        \toprule
        \textbf{Source type} & \textbf{Nutrition source} & \textbf{Min detail level} & \textbf{Proof to attach} & \textbf{Default quality check} & \textbf{Common pitfalls} \\
        \midrule
        Packaged food & per serving / per pack & 2 & label photo or clear OCR & auto checks + quick human review & duplicate SKUs; old labels \\
        \midrule
        Chain restaurant & per serving / menu size & 2 & screenshot/URL with item marked & OCR match + human confirm & menu changes; seasonal items \\
        \midrule
        Homemade / street & estimate / per 100g & 3 & short note or receipt (if any) & human-led review + model assist & portion bias; privacy; mixed plates \\
        \bottomrule
    \end{tabularx}
\end{table}

\end{document}